\documentclass[11pt,a4paper]{article}
\usepackage[hyperref]{acl2018}
\usepackage{times}
\usepackage{latexsym}
\usepackage{booktabs}
\usepackage{comment}
\usepackage{url}
\usepackage{graphicx}
\usepackage{amsfonts}
\usepackage{amsmath}
\usepackage[colorinlistoftodos]{todonotes}

\aclfinalcopy 
\def\aclpaperid{1551} 

\title{A Co-Matching Model for Multi-choice Reading Comprehension}

\author{Shuohang Wang$^1$, Mo Yu$^2$, Shiyu Chang$^2$,  Jing Jiang$^1$ \\
$^1${School of Information System, Singapore Management University} ~~~ $^2${IBM Research}\\
{\texttt {\{shwang.2014,jingjiang\}@smu.edu.sg} } \\
{\texttt {yum@us.ibm.com},  } {\texttt {shiyu.chang@ibm.com} } \\
}
\date{}

\begin{document}
\maketitle
\begin{abstract}
Multi-choice reading comprehension is a challenging task, which involves the matching between a passage and a question-answer pair.
This paper proposes a new \emph{co-matching} approach to this problem, which jointly models whether a passage can match both a question and a candidate answer.
Experimental results on the RACE dataset demonstrate that our approach achieves state-of-the-art performance.  
\end{abstract}

\section{Introduction}
\label{sec:intro}
Enabling machines to understand natural language text is arguably the ultimate goal of natural language processing, and the task of machine reading comprehension is an intermediate step towards this ultimate goal~\cite{richardson2013mctest,hermann:nips2015,hill2015goldilocks,rajpurkar2016squad,nguyen2016ms}.  
Recently, \newcite{lai2017race} released a new multi-choice machine comprehension dataset called RACE 
that was extracted from middle and high school English examinations in China.  Figure~\ref{tbl:example} shows an example passage and two related questions from RACE.  The key difference between RACE and previously released machine comprehension datasets (\emph{e.g.}, the CNN/Daily Mail dataset~\cite{hermann:nips2015} and SQuAD~\cite{rajpurkar2016squad}) is that the answers in RACE often cannot be directly extracted from the given passages, as illustrated by the two example questions (Q1 \& Q2) in Figure~\ref{tbl:example}.  Thus, answering these questions is more challenging and requires more inferences.  

Previous approaches to machine comprehension are usually based on pairwise sequence matching, where either the passage is matched against the sequence that concatenates both the question and a candidate answer~\cite{yin2016attention}, or the passage is matched against the question alone followed by a second step of selecting an answer using the matching result of the first step~\cite{lai2017race,Haichao18:aaai}.  However, these approaches may not be suitable for multi-choice 
reading comprehension
since questions and answers are often equally important.  
Matching the passage only against the question may not be meaningful and may lead to loss of information from the original passage, as we can see from the first example question in Figure~\ref{tbl:example}.  
On the other hand, concatenating the question and the answer into a single sequence for matching may not work, either, due to the loss of interaction information between a question and an answer.
As illustrated by Q2 in Figure~\ref{tbl:example}, the model may need to recognize what ``he'' and ``it'' in candidate answer~(c) refer to in the question, in order to select~(c) as the correct answer.  This observation of the RACE dataset shows that we face a new challenge of matching sequence triplets (\emph{i.e.}, passage, question and answer) instead of pairwise matching.

\begin{table*}[t]
\centering

\begin{tabular}{ll}
\toprule
\multicolumn{2}{p{16cm}}{Passage: \textit{My father wasn't a king, he was a taxi driver, but I am a prince-Prince Renato II, of the country Pontinha , an island fort on Funchal harbour.  In 1903, the king of Portugal sold the land to a wealthy British family, the Blandys, who make Madeira wine. Fourteen years ago the family decided to sell it for just EUR25,000, but nobody wanted to buy it either. I met Blandy at a party and he asked if I'd like to buy the island. Of course I said yes, but I had no money-I was just an art teacher. I tried to find some business partners, who all thought I was crazy. So I sold some of my possessions, put my savings together and bought it. Of course, my family and my friends-all thought I was mad ... If l want to have a national flag, it could be blue today, red tomorrow. ... My family sometimes drops by, and other people come every day because the country is free for tourists to visit ...}} \\
\midrule
Q1: \textit{Which statement of the following is true?} \hspace{2cm} 
& Q2: \textit{How did the author get the island?}  \\
a. \textit{The author made his living by driving.} & a. \textit{It was a present from Blandy.}\\
b. \textit{The author's wife supported to buy the island.} & b. \textit{The king sold it to him.} \\
c. \textit{Blue and red are the main colors of his national flag.} &  \textbf{c. \textit{He bought it from Blandy.}}\\
\textbf{d. \textit{People can travel around the island free of charge.}} &            d. \textit{He inherited from his father.}  \\
\bottomrule
\end{tabular}

\caption{An example passage and two related multi-choice questions. The ground-truth answers are in \textbf{bold}.}
\label{tbl:example}
\end{table*}

In this paper, we propose a new model to match a question-answer pair to a given passage.  Our \emph{co-matching} approach explicitly treats the question and the candidate answer as two sequences and jointly matches them to the given passage.  Specifically, for each position in the passage, we compute two attention-weighted vectors, where one is from the question and the other from the candidate answer.   
Then, two matching representations are constructed: the first one matches the passage with the question while the second one matches the passage with the candidate answer.   
These two newly constructed matching representations together form a \emph{co-matching state}.  Intuitively, it encodes the locational information of the question and the candidate answer matched to a specific context of the passage.  Finally, we apply a hierarchical LSTM~\cite{tang2015document} over 
the sequence of co-matching states at different positions of the passage.
Information is aggregated from word-level to sentence-level and then from sentence-level to document-level.  In this way, our model can better deal with the questions that require evidence scattered in different sentences in the passage.  Our model improves the state-of-the-art model by 3 percentage on the RACE dataset.
Our code will be released under \url{https://github.com/shuohangwang/comatch}.

\section{Model}
\label{sec:model}
\begin{figure*}[t]
\centering
\includegraphics[width=6in]{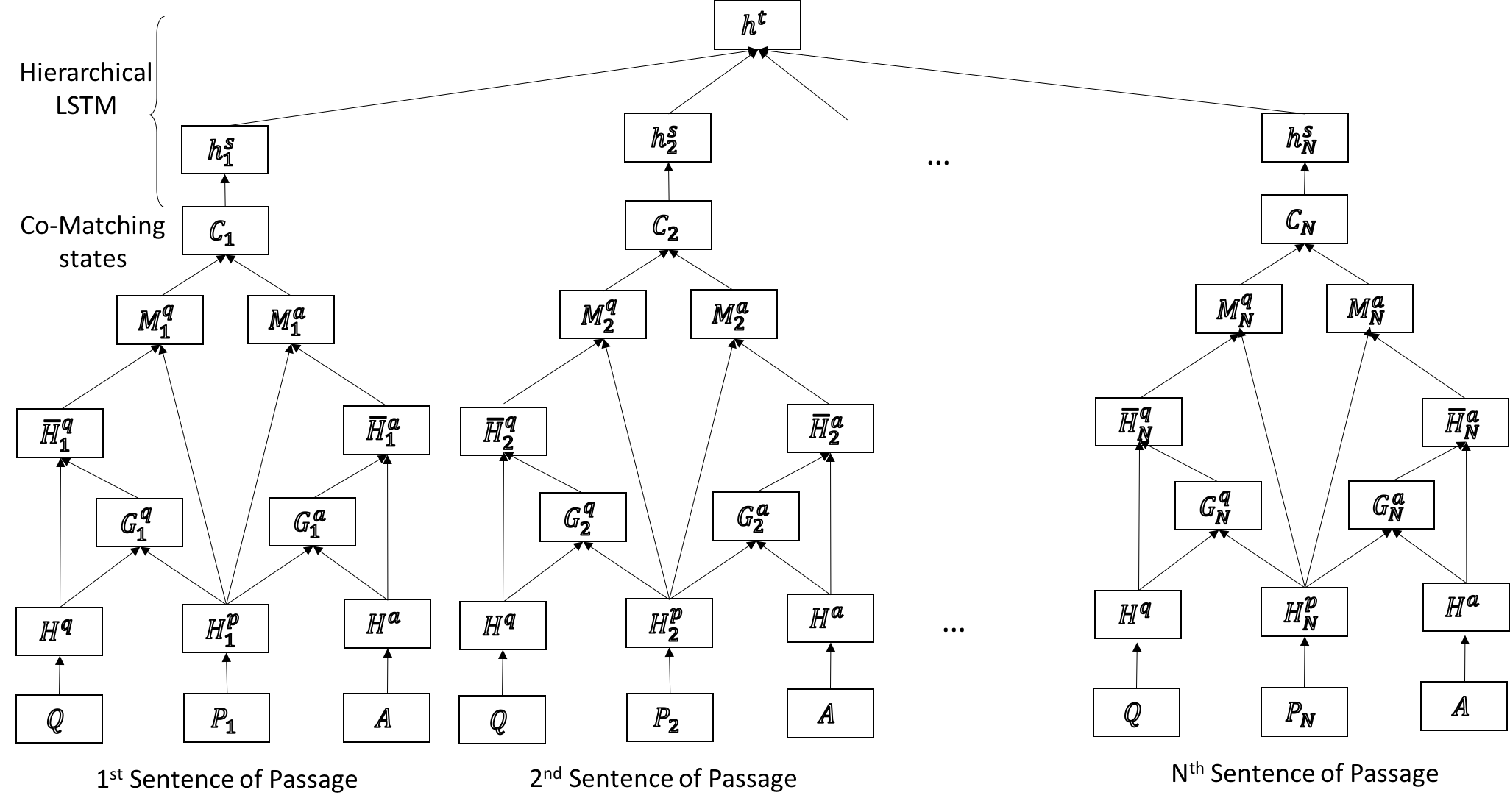}
\label{fig:model}
\caption{An overview of the model that builds a matching representation for a triplet $\{ \mathbf{P},\mathbf{Q},\mathbf{A}\}$ (\emph{i.e.}, passage, question and candidate answer).}
\end{figure*}

For the task of multi-choice reading comprehension, the machine is given a passage, a question and a set of candidate answers.  The goal is to select the correct answer from the candidates.  Let us use $\mathbf{P}\in \mathbb{R}^{d\times P}$, $\mathbf{Q}\in \mathbb{R}^{d\times Q}$ and $\mathbf{A}\in \mathbb{R}^{d \times A}$ to represent the passage, the question and a candidate answer, respectively, where each word in each sequence is represented by an embedding vector. $d$ is the dimensionality of the embeddings, and $P$, $Q$, and $A$ are the lengths of these 
sequences.

Overall our model works as follows.  For each candidate answer, our model constructs a vector that represents the matching of $\mathbf{P}$ with both $\mathbf{Q}$ and $\mathbf{A}$.  The vectors of all candidate answers are then used for answer selection.  Because we simultaneously match $\mathbf{P}$ with $\mathbf{Q}$ and $\mathbf{A}$, we call this a \emph{co-matching} model.  In Section~\ref{subsec:co-matching} we introduce the word-level co-matching mechanism.  Then in Section~\ref{subsec:aggregation} we introduce a hierarchical aggregation process.  Finally in Section~\ref{subsec:objective} we present the objective function.  An overview of our co-matching model is shown in Figure~\ref{fig:model}.

\subsection{Co-matching}
\label{subsec:co-matching}
The co-matching part of our model aims to match the passage with the question and the candidate answer at the word-level. Inspired by some previous work~\cite{wang2015learning:NAACL2016, trischler2016parallel}, we first use bi-directional LSTMs~\cite{hochreiter1997long} to pre-process the sequences as follows:
\begin{eqnarray}
\nonumber
&\mathbf{H}^{\text{p}} = \text{Bi-LSTM}(\mathbf{P}), \mathbf{H}^{\text{q}} = \text{Bi-LSTM}(\mathbf{Q}), \\
&\mathbf{H}^{\text{a}} = \text{Bi-LSTM}(\mathbf{A}),
\label{eqn:pre}
\end{eqnarray}
where $\mathbf{H}^{\text{p}}\in \mathbb{R}^{l\times P}$, $\mathbf{H}^{\text{q}}\in \mathbb{R}^{l\times Q}$  and $\mathbf{H}^{\text{a}} \in \mathbb{R}^{l\times A}$ are the sequences of hidden states generated by the bi-directional LSTMs.  We then make use of the attention mechanism to match each state in the passage to an aggregated representation of the question and the candidate answer.  The attention vectors are computed as follows:
\begin{eqnarray}
\nonumber
\mathbf{G}^{\text{q}} & = & \text{SoftMax}\left( (\mathbf{W}^{\text{g}} \mathbf{H}^{\text{q}}+\mathbf{b}^\text{g}\otimes \mathbf{e}_Q)^\text{T} \mathbf{H}^{\text{p}} \right), \\
\nonumber
\mathbf{G}^{\text{a}} & = & \text{SoftMax}\left( (\mathbf{W}^{\text{g}} \mathbf{H}^{\text{a}}+\mathbf{b}^\text{g}\otimes \mathbf{e}_Q)^\text{T} \mathbf{H}^{\text{p}} \right), \\
\nonumber
\overline{\mathbf{H}}^{\text{q}} & = & \mathbf{H}^{\text{q}}\mathbf{G}^{\text{q}}, \\
\overline{\mathbf{H}}^{\text{a}} & = & \mathbf{H}^{\text{a}}\mathbf{G}^{\text{a}},
\label{eqn:alpha}
\end{eqnarray}
where $\mathbf{W}^{\text{g}}\in \mathbb{R}^{l\times l}$ and $\mathbf{b}^{\text{g}}\in \mathbb{R}^{l}$ are the parameters to learn. $e_Q\in \mathbb{R}^{\text{Q}}$ is a vector of all 1s and it is used to repeat the bias vector into the matrix. $\mathbf{G}^{\text{q}}\in \mathbb{R}^{Q\times P}$ and $\mathbf{G}^{\text{a}}\in \mathbb{R}^{A\times P}$ are the attention weights assigned to the different hidden states in the question and the candidate answer sequences, respectively. $\overline{\mathbf{H}}^{\text{q}}\in \mathbb{R}^{l\times P}$ is the weighted sum of all the question hidden states and it represents how the question can be aligned to each hidden state in the passage.  So is $\overline{\mathbf{H}}^{\text{a}}\in \mathbb{R}^{l\times P}$.  Finally we can co-match the passage states with the question and the candidate answer as follows:
\begin{eqnarray}
\nonumber
\mathbf{M}^{\text{q}} &=& \text{ReLU}\left( \mathbf{W}^{\text{m}} \begin{bmatrix} \overline{\mathbf{H}}^{\text{q}}\ominus \mathbf{H}^{\text{p}}\\ \overline{\mathbf{H}}^{\text{q}}\otimes \mathbf{H}^{\text{p}}  \end{bmatrix} + \mathbf{b}^{\text{m}}\right), \\
\nonumber
\mathbf{M}^{\text{a}} &=& \text{ReLU}\left( \mathbf{W}^{\text{m}} \begin{bmatrix} \overline{\mathbf{H}}^{\text{a}}\ominus \mathbf{H}^{\text{p}}\\ \overline{\mathbf{H}}^{\text{a}}\otimes \mathbf{H}^{\text{p}}  \end{bmatrix} + \mathbf{b}^{\text{m}}\right), \\
\mathbf{C} &=& \begin{bmatrix}\mathbf{M}^{\text{q}} \\ \mathbf{M}^{\text{a}}  \end{bmatrix}, 
\label{eqn:match}
\end{eqnarray}
where $\mathbf{W}^{\text{g}}\in \mathbb{R}^{l\times 2l}$ and $\mathbf{b}^{\text{g}}\in \mathbb{R}^{l}$ are the parameters to learn. 
$\begin{bmatrix} \cdot \\ \cdot \end{bmatrix}$ is the column-wise concatenation of two matrices, and $\cdot\ominus \cdot$ and $\cdot\otimes \cdot$ are the element-wise subtraction and multiplication between two matrices, which are used to build better matching representations~\cite{tai2015improved,wang2016compare}. 
$\mathbf{M}^{\text{q}}\in \mathbb{R}^{l\times P}$ represents the matching between the hidden states of the passage and the corresponding attention-weighted representations of the question. 
Similarly, we match the passage with the candidate answer and represent the matching results using $\mathbf{M}^{\text{a}}\in \mathbb{R}^{l\times P}$.
Finally $C\in \mathbb{R}^{2l\times P}$ is the concatenation of $\mathbf{M}^{\text{q}}\in \mathbb{R}^{l\times P}$ and $\mathbf{M}^{\text{a}}\in \mathbb{R}^{l\times P}$ and represents how each passage state can be matched with the question and the candidate answer. 
We refer to $\mathbf{c}\in \mathbb{R}^{2l}$, which is a single column of $\mathbf{C}$, as a \emph{co-matching state} that concurrently matches a passage state with both the question and the candidate answer.


\subsection{Hierarchical Aggregation}
\label{subsec:aggregation}

In order to capture the sentence structure
of the passage, we further modify the model presented earlier and build a hierarchical LSTM~\cite{tang2015document} on top of the co-matching states. 
Specifically, we first split the passage into sentences
and we use $\mathbf{P}_1,\mathbf{P}_2, \ldots, \mathbf{P}_N$ to represent these sentences, where $N$ is the number of sentences in the passage. 
For each triplet $\{\mathbf{P}_n, \mathbf{Q}, \mathbf{A}\}, n \in [1, N]$, we can get the co-matching states $\mathbf{C}_n$ through Eqn.~(\ref{eqn:pre}-\ref{eqn:match}). 
Then we build a bi-directional LSTM followed by max pooling on top of the co-matching states of each sentence as follows:
\begin{eqnarray}
\mathbf{h}^{\text{s}}_n & = & \text{MaxPooling}\left( \text{Bi-LSTM} \left( \mathbf{C}_n \right) \right),
\end{eqnarray}
where the function $\text{MaxPooling}(\cdot)$ is the row-wise max pooling operation. $\mathbf{h}^{\text{s}}_n \in \mathbb{R}^l, n\in [1,N]$ is the sentence-level aggregation of the co-matching states.  
All these representations will be further integrated by another Bi-LSTM to get the final triplet matching representation.
\begin{eqnarray}
\nonumber
\mathbf{H}^{\text{s}} &=& [\mathbf{h}^s_1;\mathbf{h}^s_2; \ldots; \mathbf{h}^s_N], \\
\mathbf{h}^{\text{t}} & = & \text{MaxPooling}\left( \text{Bi-LSTM} \left( \mathbf{H}^{\text{s}} \right) \right),
\label{eqn:rep}
\end{eqnarray}
where $\mathbf{H}^{\text{s}}\in \mathbb{R}^{l\times N}$ is the concatenation of all the sentence-level representations and it is the input of a higher level LSTM. $\mathbf{h}^{\text{t}}\in \mathbb{R}^l$ is the final output of the matching between the sequences of the passage, the question and the candidate answer.

\subsection{Objective function}
\label{subsec:objective}

For each candidate answer $\mathbf{A}_i$, we can build its matching representation $\mathbf{h}^t_i \in \mathbb{R}^l$ with the question and the passage through Eqn.~(\ref{eqn:rep}). 
Our loss function is computed as follows:
\begin{equation}
L(\mathbf{A}_i |\mathbf{P},\mathbf{Q}) = -\log \frac{\exp (\mathbf{w}^T \mathbf{h}^t_i)}{\sum_{j=1}^4 \exp (\mathbf{w}^T \mathbf{h}^t_j)},
\end{equation}
where $\mathbf{w}\in \mathbb{R}^l$ is a parameter to learn.

\section{Experiment}
\label{sec:exp}

\begin{table}[]
\centering
\small
\begin{tabular}{lccc}
\toprule
\multicolumn{1}{c}{} & \bf RACE-M & \bf RACE-H               & \bf RACE      \\
\midrule
Random               & 24.6   & 25.0                 & 24.9      \\
Sliding Window       & 37.3   & 30.4                 & 32.2      \\
Stanford AR          & 44.2   & 43.0                 & 43.3      \\
GA                   & 43.7   & 44.2                 & 44.1      \\
ElimiNet 			 & -	  & -					 & 44.7		 \\
HAF		             & 45.3   & 47.9                 & 47.2      \\
MUSIC                & 51.5   & 45.7                 & 47.4      \\ 
\midrule
Hier-Co-Matching & \textbf{55.8}$^{*}$ & \textbf{48.2}$^{*}$ & \textbf{50.4}$^{*}$ \\
\quad - Hier-Aggregation              & 54.2   & 46.2                     & 48.5 \\
\quad - Co-Matching          & 50.7   & 45.6                 & 46.4 \\
\midrule
Turkers              & 85.1   & 69.4                 & 73.3      \\
Ceiling   & 95.4   & 94.2                 & 94.5 \\
\bottomrule
\end{tabular}
\caption{Experiment Results. $^{*}$ means it's significant to the models ablating either the hierarchical aggregation or co-matching state.}
\label{tbl:res}
\end{table}

To evaluate the effectiveness of our hierarchical co-matching model, we use the RACE dataset~\cite{lai2017race}, which consists of two subsets:
RACE-M comes from middle school examinations while RACE-H comes from high school examinations. 
RACE is the combination of the two.

We compare our model with a number of baseline models.  We also compare with two variants of our model for an ablation study.

\paragraph{Comparison with Baselines} We compare our model with the following baselines:

$\bullet$~\textbf{Sliding Window} based method~\cite{richardson2013mctest} computes the matching score based on the sum of the tf-idf values of the matched words between the question-answer pair and each sub-passage with a fixed a window size.  

$\bullet$~\textbf{Stanford Attentive Reader (AR)}~\cite{chen2016thorough} first builds a question-related passage representation through attention mechanism and then compares it with each candidate answer representation to get the answer probabilities. 

$\bullet$~\textbf{GA}~\cite{dhingra2016gated} uses gated attention mechanism with multiple hops to extract the question-related information of the passage and compares it with candidate answers. 

$\bullet$~\textbf{ElimiNet}~\cite{eliminet} tries to first eliminate the most irrelevant choices and then select the best answer. 

$\bullet$~\textbf{HAF}~\cite{Haichao18:aaai} considers not only the matching between the three sequences, namely, passage, question and candidate answer, but also the matching between the candidate answers.   

$\bullet$~\textbf{MUSIC} \cite{xu2017towards} integrates different sequence matching strategies into the model and also adds a unit of multi-step reasoning for selecting the answer.  

Besides, we also report the following two results as reference points: \textbf{Turkers} is the performance of Amazon Turkers on a randomly sampled subset of the RACE test set. \textbf{Ceiling} is the percentage of the unambiguous questions with a correct answer in a subset of the test set.  

The performance of our model together with the baselines are shown in Table~\ref{tbl:res}. We can see that our proposed complete model, \textbf{Hier-Co-Matching}, achieved the best performance among all the public results. Still, there is a huge gap between the best machine reading performance and the human performance, showing the great potential for further research.

\paragraph{Ablation Study} Moreover, we conduct an ablation study of our model architecture. 
In this study, we are mainly interested in the contribution of each component introduced in this work to our final results. We studied two key factors: (1) the co-matching module and (2) the hierarchical aggregation approach.  We observed a 4 percentage performance decrease by replacing the co-matching module with a single matching state (\emph{i.e.}, only $\mathbf{M}^a$ in Eqn~(\ref{eqn:match})) by directly concatenating the question with each candidate answer~\cite{yin2016attention}. We also observe about 2 percentage decrease when we treat the passage as a plain sequence, and run a two-layer LSTM (to ensure the numbers of parameters are comparable) over the whole passage instead of the hierarchical LSTM.

\paragraph{Question Type Analysis} We also conducted an analysis on what types of questions our model can handle better. 
We find that our model obtains similar performance on the ``wh'' questions such as ``why,'' ``what,'' ``when'' and ``where'' questions, on which the performance is usually around 50\%.
We also check statement-justification questions with the keyword ``true'' (\emph{e.g.}, ``Which of the following statements is true"), negation questions with the keyword ``not'' (\emph{e.g.}, ``which of the following is not true''),  and summarization questions with the keyword ``title'' (\emph{e.g.}, ``what is the best title for the passage?''), and their performance is 51\%, 52\% and 48\%, respectively. 
We can see that the performance of our model on different types of questions in the RACE dataset is quite similar.
However, our model is only based on word-level matching and may not have the ability of reasoning. 
In order to answer questions that require summarization, inference or reasoning, we still need to further explore the dataset and improve the model.
Finally, we further compared our model to the baseline, which concatenates the question with each candidate answer, and our model can achieve better performance on different types of questions. For example, on the subset of the questions with pronouns, our model can achieve better accuracy of 49.8\% than 47.9\%. Similarly, on statement-justification questions with the keyword ``true'', our model could achieve better accuracy of 51\% than 47\%.

\section{Conclusions}
In this paper, we proposed a co-matching model for multi-choice reading comprehension.
The model consists of a co-matching component and a hierarchical aggregation component.
We showed that our model could achieve state-of-the-art performance on the RACE dataset.
In the future, we will adapt the idea of co-matching and hierarchical aggregation to the standard open-domain QA setting for answer candidate reranking \cite{wang2017evidence}.
We will also further study how to explicitly model inference and reasoning on the RACE dataset.

\section{Acknowledgement}
This work was partially supported by DSO grant DSOCL15223.

\bibliography{acl2018}
\bibliographystyle{acl_natbib}

\end{document}